\documentclass[letterpaper, 10 pt, conference]{ieeeconf}  

\IEEEoverridecommandlockouts                              

\usepackage{cuted}
\usepackage{acronym}
\usepackage{graphics} 
\usepackage{float} 
\usepackage{stfloats} 
\usepackage{array}
\usepackage{epsfig} 
\usepackage{bm}
\usepackage{amsmath} 
\usepackage{amssymb}  
\usepackage{cite}
\usepackage{booktabs}
\usepackage{algorithm}
\usepackage{algorithmicx}
\usepackage{algpseudocode}
\usepackage{multirow}
\usepackage{relsize}
\usepackage{caption}
\usepackage{siunitx}
\usepackage{todonotes}
\usepackage{hyperref}

\newif\ifanon
\anonfalse  

\title{\LARGE \bf Visual Cue Guided Video Planning for Generalizable Robot Navigation
}

\ifanon
\author{%
 \\}
\else
\author{%
Hojin Lee$^{1}$,
Sizhe Lester Li$^{2}$,
Maximilian Hilger$^{1}$,
Susie Lu$^{2}$,
Achim J. Lilienthal$^{1,3}$,\\
Vincent Sitzmann$^{2,\dagger}$,
and Daniel A Duecker$^{1,\dagger}$%
\thanks{$^{\dagger}$Equal advising.}%
\thanks{$^{1}$
Munich Institute of Robotics and Machine Intelligence (MIRMI),
Technical University of Munich, 80797 Munich, Germany.}%
\thanks{$^{2}$
Computer Science and Artificial Intelligence Laboratory (CSAIL),
Massachusetts Institute of Technology, Cambridge, MA 02139, USA.}%
\thanks{$^{3}$Centre for Applied Autonomous Sensor Systems (AASS), Örebro University, 72081 Örebro, Sweden.}%
\thanks{This work was supported by the Bavarian State Ministry of Science
and the Arts (StMWK) through the MIT--TUM Collaboration
(grant no.~151223031).}%
\thanks{Vincent Sitzmann and Sizhe Lester Li were supported by the National Science Foundation under Grant No. EEC 2330040 and the CAREER program under 2543631.}
}
\fi

\begin{document}

\acrodef{don}[DoN]{Degree of Narrowness}
\acrodef{ne}[NE]{navigation error}
\acrodef{os}[OS]{oracle success}
\acrodef{spl}[SPL]{success weighted by path length}
\acrodef{sr}[SR]{success rate}
\acrodef{tl}[TL]{trajectory length}
\acrodef{mdbi}[MDBI]{Mean Distance Between Interventions}
\acrodef{mtbi}[MTBI]{Mean Time Between Interventions}

\acrodef{bev}[BEV]{Bird’s-Eye View}
\acrodef{idm}[IDM]{Inverse-Dynamics Model}
\acrodef{lora}[LoRA]{Low-Rank Adaptation}
\acrodef{mlp}[MLP]{Multi Layer Perceptron}
\acrodef{vae}[VAE]{Variational Autoencoder}
\acrodef{vla}[VLA]{Vision-Language-Action Model}
\acrodef{vlm}[VLM]{Vision-Language Model}
\acrodef{vln}[VLN]{Vision-Language Navigation}

\maketitle

\thispagestyle{empty}
\pagestyle{empty}

\begin{strip}
\ifanon
\else
\vspace*{-7.2\baselineskip}
\fi
  \centering
  \includegraphics[width=\textwidth]{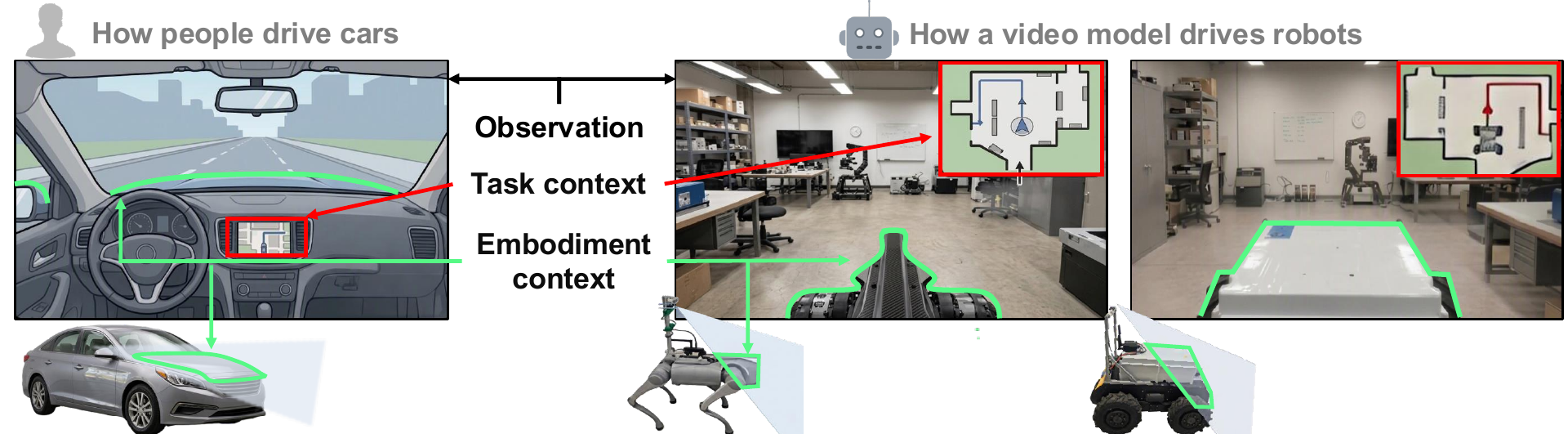}
  \captionof{figure}{\textbf{Visual cue guided video planning for generalizable robot navigation (CueNav)}. Similar to human driving, a video planner can benefit from task and embodiment context expressed through the visual observation. We illustrate these forms of context using a map of the surroundings and a partial view of the robot body.
}
  \label{fig:teaser}
\end{strip}


\renewcommand{\thefootnote}{\fnsymbol{footnote}}
\begin{abstract}
Generative video models can serve as a promising backbone for robot navigation by predicting future observations as video plans. Recent approaches often condition video planning on short-horizon guidance and recover geometric waypoints through scene reconstruction, leaving longer-horizon planning and precise video-to-action translation less explored. We present \textit{CueNav}, a video model-based navigation framework combining visual cue guided video planning with an embodiment-specific \ac{idm}. As visual cues, we use a \ac{bev} map to convey global task context and retain part of the robot body in the egocentric observation to expose embodiment context. These cues guide the video planner, while the \ac{idm} translates dense flow fields extracted from the video plan into robot actions. With the visual cue encoding global task context, CueNav achieves nearly $2\times$ higher success in maze navigation than planning without the cue. The body-aware view with the \ac{idm} enables precise navigation with \qty{70}{\percent} success in a narrow passage where comparison methods largely fail to complete the task. We further demonstrate zero-shot semantic-conditioned navigation and deployment of the same video planner across different robot platforms. Our results show that visual cue-guided video planning with embodiment-specific action grounding paves the way toward a generalizable navigation framework for longer-horizon planning and embodiment-aware control. Additional results and code are available on our project website\footnote[1]{Project website: \url{https://cuenav.github.io}}.
\end{abstract}
\renewcommand{\thefootnote}{\arabic{footnote}}

\section{INTRODUCTION}
General-purpose robot navigation requires robots to operate in unseen environments and respond to diverse task specifications. 
Recent \ac{vln} methods have made substantial progress toward this goal by building on large pretrained \acp{vlm}, which are finetuned into \acp{vla}~\cite{wei2026ground}. 
These models provide strong semantic reasoning and can support a range of navigation tasks within a unified framework~\cite{liu2025x,gong2026abot,sridhar2024nomad}. 
However, \acp{vlm} encounter two problems when being deployed in \ac{vln}:
They require large amounts of training data to adapt to the robot domain, and fine-tuning on robot-action data limits their generality across tasks, environments, and robot platforms~\cite{grover2025enhancing}.

Video models offer a promising alternative through broad priors from internet-scale video data~\cite{chen2026imaginav}. 
Given recent observations and task instructions, they can predict how the scene should evolve as the task is accomplished. 
This allows navigation behavior to exploit semantic and motion priors learned from large-scale video data without directly committing the planner to the action space of a specific robot. 
Recent works have shown that generated future observations can provide useful navigation guidance and generalize to novel scenes and goals~\cite{huang2026navdreamer,chen2026imaginav}.

Still, several challenges remain for
video models in navigation:
\begin{itemize}
    \item Existing approaches require granular instructions, e.g., turning left or right \cite{chen2026imaginav}. This requirement for local directional guidance does not fully utilize the capabilities of 
    video models.
    \item Existing video planners do not account for the robot's physical characteristics~\cite {liu2026imagineuav}. Knowing the robot's size and kinematics is required for precise and accurate planning in constrained spaces.
    \item Existing methods map video plans to actions via waypoints, which require a dedicated downstream controller to convert them into control commands that the robot can feasibly execute~\cite{huang2026navdreamer,chen2026imaginav,liu2026imagineuav}. As a result, the generated visual motion is not directly grounded in the robot's kinematics and dynamics.
\end{itemize}

To address these challenges, we introduce CueNav, a video-based navigation framework that guides video planning through task and embodiment visual cues and grounds the resulting visual motion into continuous robot control. 
Our approach is built around two design principles:
\begin{itemize}
\item \textbf{Visual cues for task and embodiment aware planning.}
We enable longer-horizon tasks and embodiment-awareness by providing visual cues to the video model.
For navigation from high-level task specifications, we embed a \ac{bev} map into the visual observation to provide global task context (See Fig.~\ref{fig:teaser}). 
For embodiment-aware navigation, we retain part of the robot body within the egocentric view, exposing cues about its geometry and spatial relationship to nearby obstacles and free space. 
\item \textbf{Translate visual motion through an \ac{idm}.}
We infer robot actions from dense flow fields in the generated video using an embodiment-specific \ac{idm}. 
The \ac{idm} captures the kinematics and dynamics of each robot, enabling precise execution while preserving a shared video planner across embodiments. 
\end{itemize}

We evaluate CueNav through a series of simulation and real-world experiments and show that (i) the video planner enables zero-shot semantic-conditioned navigation; (ii) global task context provided through the observation enables planning beyond local directional guidance and generalization to larger, unseen environments; (iii) embodiment-visible observations and the embodiment-specific flow-based \ac{idm} improve precise navigation in geometrically constrained spaces; and (iv) the same video planner can be deployed across wheeled and legged robot platforms using embodiment-specific \acp{idm}. Together, these results suggest that visual cue-guided video planning, paired with a flow-based \ac{idm}, offers a promising direction for general robot navigation using video models.

\section{RELATED WORK}

\subsection{Foundation Models for Vision-Language Navigation}
Recent methods build on large pretrained \acp{vla} to support multiple navigation tasks within a unified framework~\cite{gong2026abot,majumdar2026robostral}. These models have demonstrated strong semantic reasoning across tasks such as instruction following~\cite{anderson2018vision}, object-goal navigation~\cite{yokoyama2024hm3d}, and person following~\cite{wang2025trackvla}. 

Despite this progress, learning robot-specific control still requires large amounts of paired observation-action data. Recent navigation foundation models are trained at correspondingly large scale, from 8M navigation samples in NavFoM~\cite{zhang2026embodied} to 30M supervised samples in ABot-N1~\cite{gong2026abot}. Since collecting comparable amounts of real-world robot data is costly, much of this data is generated in simulation or reconstructed environments, which can introduce sim-to-real gaps and couple the learned policy to particular action spaces and embodiments.

Video models offer a promising alternative by learning semantic and motion priors from internet-scale videos without requiring robot action labels. Rather than predicting robot actions directly, they allow navigation behavior to be planned in visual space, while robot-specific control can be learned separately.

\subsection{Video Models for Robot Navigation}
Recent works use video models as visual planners for robot navigation, differing mainly in how future behavior is specified and translated into robot motion. ImagiNav~\cite{chen2026imaginav} and ImagineUAV~\cite{liu2026imagineuav} condition future visual trajectories on navigation instructions, but their evaluations largely rely on local motion guidance, leaving task-level navigation without explicit directional cues less explored. 
NavDreamer~\cite{huang2026navdreamer}, ImagiNav~\cite{chen2026imaginav}, and ImagineUAV~\cite{liu2026imagineuav} recover camera motion or geometric waypoints from generated video for downstream control. This introduces additional geometric estimation and tracking stages, while the visual planner itself does not explicitly account for the executing embodiment, although robot size, shape, and motion capabilities affect which behaviors are physically feasible. 
DreamToNav~\cite{serpiva2026dreamtonav} generates videos in a third-person perspective, containing the robot environment. However, to recover the motion of the robot, both camera pose and robot pose in the camera frame need to be estimated, which complicates the motion reconstruction.
SparseVideoNav~\cite{zhang2026sparse} avoids explicit geometric reconstruction by predicting actions from generated future visual features. However, this still requires paired visual and action supervision, retaining limitations similar to \acp{vla}. 

In contrast, CueNav uses global task context to support navigation from high-level objectives without explicit local directional guidance, while an embodiment-specific flow-based \ac{idm} grounds generated visual motion directly into continuous robot commands without explicit geometric reconstruction or joint video-action training.

\begin{figure*}[t]
  \centering
  \includegraphics[width=0.91\textwidth]{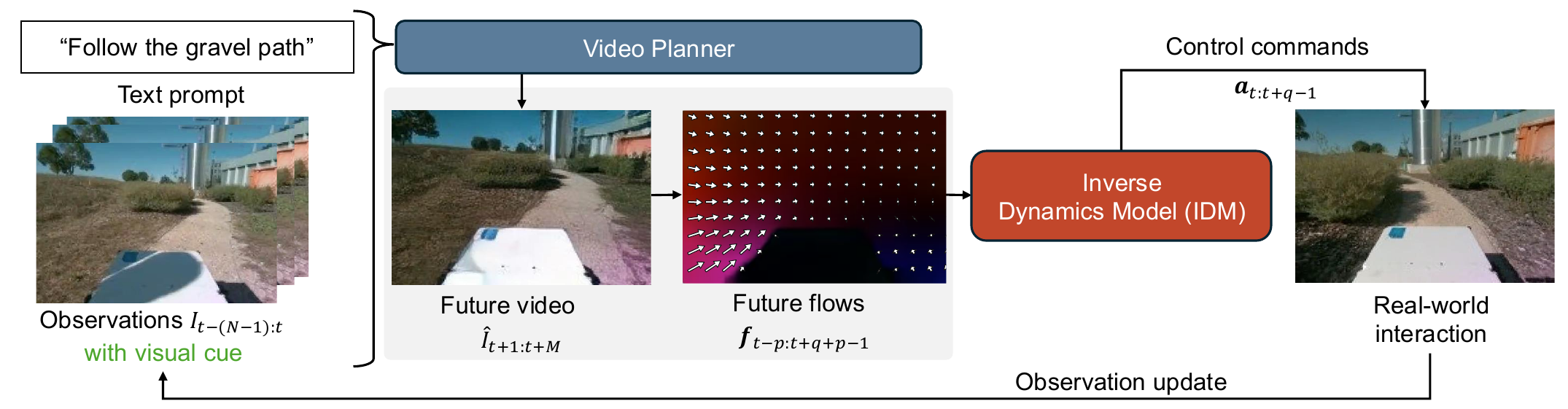}
 \caption{Overview of CueNav. Given recent visual observations with embodiment or global task context and a text prompt, the video planner predicts short-horizon future observations. The resulting flow fields from the predicted video are mapped to continuous robot actions by the \ac{idm}, followed by closed-loop replanning from new observations.}
\label{fig:overview}\vspace{-4mm}
\end{figure*}

\section{Main Method}
\subsection{Problem Formulation}
A video model $\mathcal{P}$ produces a short-horizon visual plan for navigation. Given a recent observation history $\mathbf{I}_{t-(N-1):t}=\{I_{t-(N-1)},\ldots,I_t\}$ and a navigation prompt $g$, the video model samples a sequence of future observations as
\begin{equation}\label{eq:video_model}
    \hat{\mathbf{I}}_{t+1:t+M}
    \sim
    \mathcal{P}(\cdot \mid \mathbf{I}_{t-(N-1):t}, g),
\end{equation}
where $N$ denotes the observation history length and $M$ is the prediction horizon. The predicted future observations encode the intended navigation motion. To execute this motion, the visual plan must ultimately be translated into a sequence of continuous control commands. Hence, this video-to-navigation control formulation raises two central questions concerning how to generate navigation-relevant visual plans and how to ground them in executable robot control.

\subsection{Visual Cue Guided Planning with a Video Model}
The video planner predicts short-horizon future observations conditioned on recent observations and a navigation prompt. A key design choice is how the visual input is constructed. To support generalizable navigation, we expose task and embodiment context directly through visual cues.

\begin{figure}[htp]
  \centering
  \includegraphics[width=0.49\textwidth]{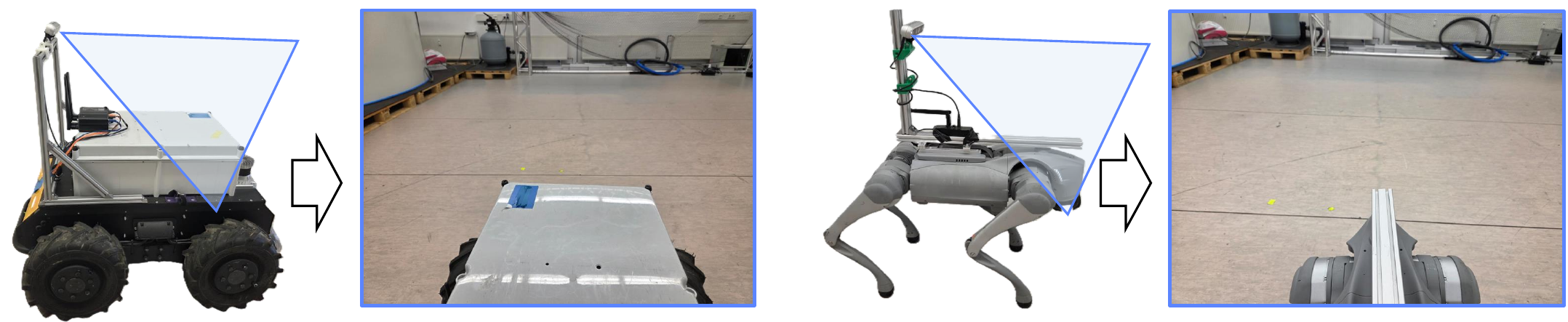}
  \caption{Embodiment-aware observation with part of the robot body visible as a visual cue.}
  \label{fig:step}\vspace{-3mm}
\end{figure}

\textbf{Visual Cue Design.}
We consider two visual cue designs. First, for navigation from high-level goal specifications, we embed a \ac{bev} map into the observation (Fig.~\ref{fig:maze_trajectory}(a)). The map contains the robot and goal locations while leaving the path between them unspecified, providing global situational awareness beyond the egocentric view. Second, for embodiment-aware navigation, we position the camera such that parts of the robot body remain visible in the egocentric observation (Fig.~\ref{fig:step}). This exposes cues about the robot’s geometry and its spatial relationship to nearby obstacles and free space, without requiring a separate embodiment representation or an explicit geometric description.

\textbf{Action-free Video Model Post-training.}
We build the video planner on the open-weight Wan2.2-5B video diffusion transformer~\cite{wan2025wan} and adapt the pretrained TI2V backbone for video-to-video prediction. We choose this model for its open availability and practical balance between model capacity and onboard computational cost, while CueNav can in principle be used with other video-model backbones. We use navigation videos only and do not require robot action labels. During post-training, each navigation video is divided into $N$ context frames and $M$ future frames, with the visual cues present in all frames. The context frames and navigation prompt condition the model. The training loss uses the same generative prediction objective as during video model pretraining and is applied only to future frames. We follow a diffusion-forcing formulation for autoregressive video prediction~\cite{chen2024diffusion}.

At deployment, the planner predicts a short-horizon visual future from the latest observation history. The predicted video frames are passed to the flow-based \ac{idm} described in the following section, which maps the predicted visual motion to continuous robot commands. After executing these commands, newly observed frames are used to update the observation history for the next planning step, enabling closed-loop receding-horizon planning. An overview of the proposed framework is shown in Fig.~\ref{fig:overview}.

\begin{figure*}[htpb]
  \centering
  \includegraphics[width=0.87\textwidth]{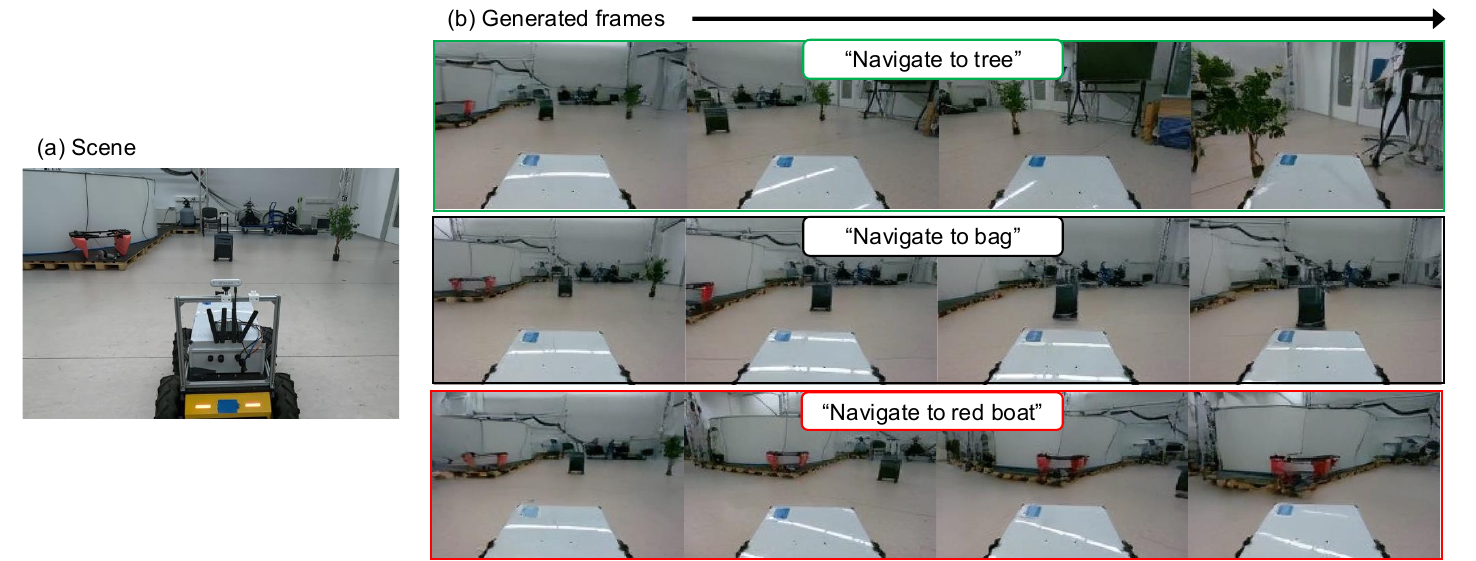}
  \caption{Zero-shot semantic-conditioned visual planning. (a) Initial scene.
(b) Generated future observations conditioned on different semantic goal prompts.}
  \label{fig:semantics}\vspace{-5mm}
\end{figure*}

\subsection{Flow-based Inverse-Dynamics Model}
The predicted video describes how the scene should evolve but does not directly specify executable robot actions, $\mathbf{a}_t\in\mathbb{R}^{n}$. We infer these actions from the predicted frames using an \ac{idm}, $\pi_{\mathrm{IDM}}$. The \ac{idm} uses dense flow fields between consecutive frames rather than the predicted RGB frames directly, inspired by the use of flow for action inference in~\cite{li2026turning}. Specifically, at time step $t$, it predicts a sequence of $q$ robot actions as
\begin{equation}\label{eq:idm}
    \mathbf{a}_{t:t+q-1}
    =
    \pi_{\mathrm{IDM}}
    \left(
    \textbf{f}_{t-p}, \ldots, \textbf{f}_{t+q+p-1}
    \right),
\end{equation}

where
$\mathbf{a}_{t:t+q-1}
=
\{\mathbf{a}_t,\ldots,\mathbf{a}_{t+q-1}\}$
and
$\textbf{f}_i\in\mathbb{R}^{H\times W\times 2}$
denotes the dense flow field computed from the frame pair at time steps
$i$ and $i+1$ of the combined observed-and-predicted sequence
$\{\mathbf{I}_{t-(N-1):t},\hat{\mathbf{I}}_{t+1:t+M}\}$. The $p\geq1$ additional flow fields at both ends of the window provide temporal context, allowing the \ac{idm} to account for system dynamics such as delayed control responses. In total, the window contains $q+2p$ flow fields, which requires $p+q\leq M$ and $p\leq N-1$. We compute the flow fields using AllTracker as an off-the-shelf flow estimator~\cite{harley2025alltracker}.

\textbf{IDM with Spatiotemporal Transformer.}
The \ac{idm} consists of a shared convolutional encoder followed by a spatiotemporal transformer. Each flow field is encoded by the convolutional encoder and adaptively average-pooled to a fixed spatial grid. The resulting spatial tokens from all flow fields are processed jointly by the transformer with 3D rotary positional embeddings over space and time. A feed-forward network maps the resulting features to the $q$ predicted actions. The \ac{idm} is trained to regress the corresponding robot actions using a mean-squared error objective.

\textbf{Embodiment-specific Training.}
We train a separate \ac{idm} for each robot platform using paired flow fields and robot actions, where the flow fields are computed from RGB observations recorded during navigation. For wheeled robots, the action space is $\mathbf{a}=(v_x,\omega_z)\in\mathbb{R}^{2}$, where $v_x$ and $\omega_z$ denote the forward linear velocity and yaw angular velocity, respectively. For legged robots, we use $\mathbf{a}=(v_x,v_y,\omega_z)\in\mathbb{R}^{3}$, additionally including the lateral linear velocity $v_y$. 

\textbf{Closed-loop Execution.}
At deployment, the flow fields from the predicted frames are passed to the \ac{idm} to obtain a sequence of robot actions. We execute $\mathbf{a}_{t:t+q-1}$ actions before acquiring new observations and querying the video planner again. This receding-horizon procedure, summarized in Algorithm~\ref{alg:cuenav}, updates the visual plan using real observations and translates the predicted visual motion into robot-specific control inputs.

\begin{algorithm}[htpb]
\caption{CueNav closed-loop navigation}
\label{alg:cuenav}
\textbf{Initialize:} video planner $\mathcal{P}$,
IDM $\pi_{\mathrm{IDM}}$, text prompt $g$,
video context length $N$, video prediction horizon $M$, action horizon $q$, and temporal padding length $p$.
\begin{algorithmic}[1]
\While{not terminated}
    \State Acquire recent observations $\mathbf{I}_{t-(N-1):t}$
    \State Sample future video frames using \eqref{eq:video_model}
    \State Compute flow fields $\textbf{f}_{t-p:t+q+p-1}$
    \State Infer robot actions using \eqref{eq:idm} and execute the actions
    \State Append the newly acquired observations
    \State $t \gets t + q$
\EndWhile
\end{algorithmic}
\end{algorithm}
\vspace{-3mm}

\section{Experiments}
To validate the main design choices of CueNav, we organize our experiments around three questions.

\textbf{Semantic-conditioned closed-loop navigation.}
Can CueNav leverage the generalization capability of the video model to enable semantic-conditioned closed-loop navigation in real environments? (Sec.~\ref{subsec:language_navigation})

\textbf{Precise embodiment-aware navigation.}
Do embodiment cues in the visual observation improve navigation in geometrically constrained spaces? (Sec.~\ref{subsec:precise_navigation})

\textbf{Planning beyond local directional guidance.}
Does global task context enable navigation beyond local directional guidance? (Sec.~\ref{subsec:maze})

\subsection{Experimental Setup}
\label{subsec:experimental_setup}

\textbf{Robotic Platforms and Data Collection.}
We evaluate CueNav on two ground robot platforms: a Clearpath Husky A300 wheeled robot and a Unitree Go2 quadruped, both controlled through their low-level velocity controllers. Both robots are equipped with a RealSense D455 camera, of which we use only RGB images. For each platform, we collect approximately two hours of navigation data in indoor and outdoor environments around industrial buildings, consisting of RGB observations paired with the executed robot actions. The video planner is post-trained using only RGB observations from both platforms, without requiring action labels. A separate \ac{idm} is trained for each platform using paired flow fields and robot actions from the corresponding robot navigation data. We use the Husky A300 for semantic-conditioned and precise embodiment-aware navigation experiments, where its non-holonomic motion makes constrained navigation particularly challenging. The Unitree Go2 is used to demonstrate cross-embodiment deployment of the shared video planner.

\textbf{Implementation Details.}
We build the video planner on the pretrained Wan2.2-5B video model~\cite{wan2025wan}, operating at a spatial resolution of $192\times128$ pixels (width $\times$ height) and a frame rate of \qty{10}{\hertz}. The pretrained \acl{vae} remains frozen during post-training, while \acl{lora}~\cite{hu2021lora} is applied to the diffusion transformer. Post-training is performed on a single AMD MI300X GPU using mixed-precision training where supported.

During inference, the model conditions on $N=21$ context frames and generates $M=16$ future frames as the visual plan. CueNav runs fully onboard on an NVIDIA Jetson Thor, where generating one visual plan and inferring the corresponding robot actions with the \ac{idm} takes approximately \qty{4}{\second}. For both robot platforms, we use $p=1$ temporal padding flow fields and execute the $q=15$ predicted actions after each video-planning step.

\textbf{Baselines and Ablations.}
We compare \textit{CueNav} against two publicly available vision-language navigation methods, \textit{StreamVLN}~\cite{wei2025streamvln} and \textit{InternVLA-N1}~\cite{wei2026ground}. We select these as strong recent \ac{vln} baselines with publicly released implementations and checkpoints. To our knowledge, there are no publicly available models and checkpoints for closely related video-model-based navigation methods. Both baselines are evaluated using their released checkpoints and egocentric observations consistent with their respective training setups. For \textit{InternVLA-N1}, when the policy requests a camera look-down action that cannot be executed by our fixed-camera platform, we instead rotate the robot to acquire a new observation. For \textit{StreamVLN}, we additionally attempted fine-tuning with our in-domain navigation data. Since this degraded performance relative to the released checkpoint, we report results using the released model without additional fine-tuning. To isolate the effect of embodiment cues, we also train a \textit{CueNav} variant using a pure egocentric observation in which the robot body is not visible, denoted as \textit{CueNav (w/o body)}. To evaluate the proposed flow-based \ac{idm}, we additionally consider \textit{CueNav (w/o IDM)}, which uses the same visual observations and video planner as \textit{CueNav} but replaces the \ac{idm} with geometry-based grounding similar to that in~\cite{chen2026imaginav}. We use MASt3R \cite{leroy2024grounding} to reconstruct the scene and recover the camera trajectory from the generated video, which is then tracked by a simple heuristic controller. All other training and inference settings are kept unchanged.

\subsection{Semantic-conditioned Goal Navigation}
\label{subsec:language_navigation}
We first evaluate whether \textit{CueNav} can leverage the generalization capability of the video planner for semantic-conditioned closed-loop navigation in real environments. The robot is given a semantic target description and must navigate to the corresponding object.

\textbf{Task Setup and Evaluation Metrics.}
We place three unseen semantic targets in the environment and prompt the robot to navigate toward each target from the same initial position. Each target is evaluated over 10 runs, resulting in 30 trials per method. 

We report \ac{sr}, \ac{spl}, final \ac{ne}, and \ac{tl}. A trial is considered successful when the robot terminates within \qty{1.5}{\meter} of the target center. \ac{sr} measures the fraction of successful trials, while \ac{spl} is defined as
$
\mathrm{SPL}
=
\frac{1}{E}
\sum_{i=1}^{E}
S_i
\frac{\ell_i}{\max(p_i,\ell_i)}
$,
where $E$ is the total number of trials, $S_i\in\{0,1\}$ indicates success, $\ell_i$ is the shortest-path distance from the start to the target, and $p_i$ is the executed path length. \ac{ne} measures the final distance between the robot and the target center, and \ac{tl} denotes the total distance traveled during the trial.

\begin{table}[htpb]
  \centering\vspace{-1mm}
  \caption{Zero-shot semantic-conditioned goal navigation performance over 30 trials per method. NE and TL are reported as mean $\pm$ standard deviation.}
  \label{tab:semantic_conditioning}
  \begin{tabular}{lcccc}
    \toprule
    Method                                       & SR $\uparrow$  & SPL $\uparrow$ & NE [m]                   & TL [m]                   \\
    \midrule
    StreamVLN~\cite{wei2025streamvln}            & 0.467          & 0.418          & $1.88 \pm 1.58$          & $3.69 \pm 0.97$          \\
    InternVLA-N1~\cite{wei2026ground}            & 0.733          & 0.665          & $1.22 \pm 1.03$          & $3.62 \pm 1.29$          \\
    CueNav (w/o body)                            & 0.900          & 0.833          & $\mathbf{0.70 \pm 0.53}$ & $3.35 \pm 0.56$          \\
    CueNav (w/o IDM)                             & 0.667          & 0.615          & $1.25\,\pm\,0.80$          & $3.31\,\pm\,0.78$      \\
    CueNav                                       & \textbf{0.933} & \textbf{0.913} & $0.84 \pm 0.71$          & $\mathbf{3.02 \pm 0.72}$ \\
    \bottomrule
  \end{tabular}\vspace{-1mm}
\end{table}

\textbf{Results.}
Table~\ref{tab:semantic_conditioning} summarizes the zero-shot semantic-conditioned navigation performance. \textit{CueNav} achieves the highest \ac{sr} and \ac{spl}. \textit{CueNav (w/o body)} remains competitive and attains the lowest final navigation error, indicating that the video planner retains strong semantic goal-conditioning capability even without an embodiment-visible observation. In contrast, even with the same video planner, \textit{CueNav (w/o IDM)}, which uses geometry-based camera-pose reconstruction, substantially reduces \ac{sr} and \ac{spl}, highlighting the importance of accurate visual-motion grounding. Overall, these results suggest that semantic-conditioned video-based navigation remains effective without an embodiment-visible observation, but depends more strongly on faithful grounding of the generated visual motion into robot actions. Fig.~\ref{fig:semantics} shows the future observations generated by \textit{CueNav} during closed-loop navigation toward different semantic targets.

\begin{figure*}[t]
  \centering
  \includegraphics[width=0.99\textwidth]{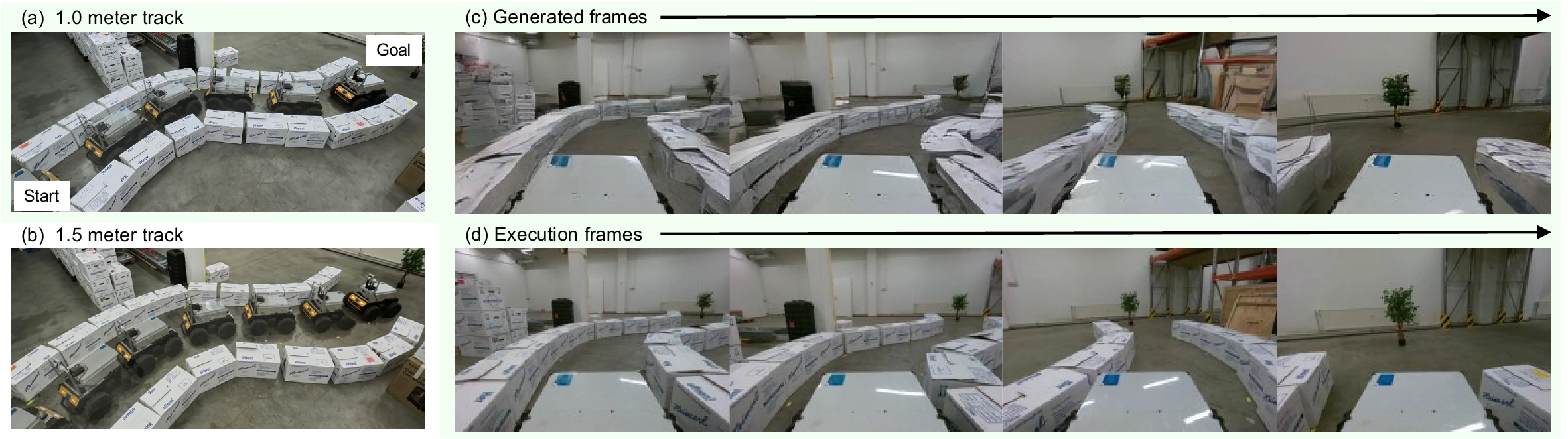}
\caption{Zero-shot precise embodiment-aware navigation through narrow tracks.
(a,b) CueNav trajectories overlaid on the \qty{1.0}{\meter}- and \qty{1.5}{\meter}-wide track scenes, respectively.
(c) Generated future observations for a representative rollout in the \qty{1.0}{\meter} track.
(d) Corresponding real observations at matched timesteps.}
  \label{fig:narrow_gap_1m}\vspace{-3mm}
\end{figure*}

\subsection{Precise Embodiment-aware Navigation}
\label{subsec:precise_navigation}
We next evaluate whether including an embodiment cue by keeping part of the robot body visible improves precise navigation in geometrically constrained spaces.

\textbf{Task Setup and Evaluation Metric.}
We construct narrow navigation tracks using boxes and require the Husky robot to traverse the constrained passage without becoming immobilized by collisions. These narrow-track environments are not included during post-training, making this a zero-shot precise-navigation task. We consider two track widths, \qty{1.5}{\meter} and \qty{1.0}{\meter}, as shown in Fig.~\ref{fig:narrow_gap_1m}. The Husky is \qty{0.7}{\meter} wide, making the \qty{1.0}{\meter} track substantially more constrained. Each condition is evaluated over 10 runs. A trial terminates when the robot reaches the end of the track or becomes immobilized upon contact with surrounding obstacles. We evaluate the maximum normalized progress reached along the track before termination, where 0 denotes the start, and 1 denotes successful completion.

\begin{figure}[htpb]
   \centering
   \includegraphics[width=0.49\textwidth]{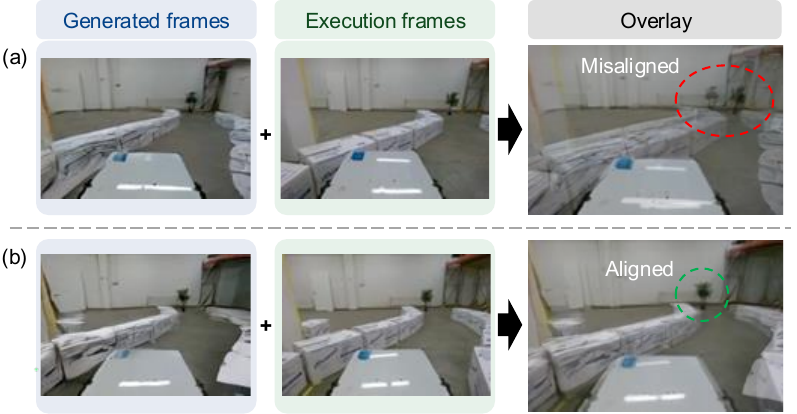}
\caption{Visual-plan realization with and without the proposed IDM.
(a) CueNav w/o IDM shows visible misalignment between generated and realized observations.
(b) CueNav maintains close alignment.}
\label{fig:precisenavigation_pathoverlay}\vspace{-5mm}
\end{figure}

\begin{figure}[htpb]
   \centering
   \includegraphics[width=0.45\textwidth]{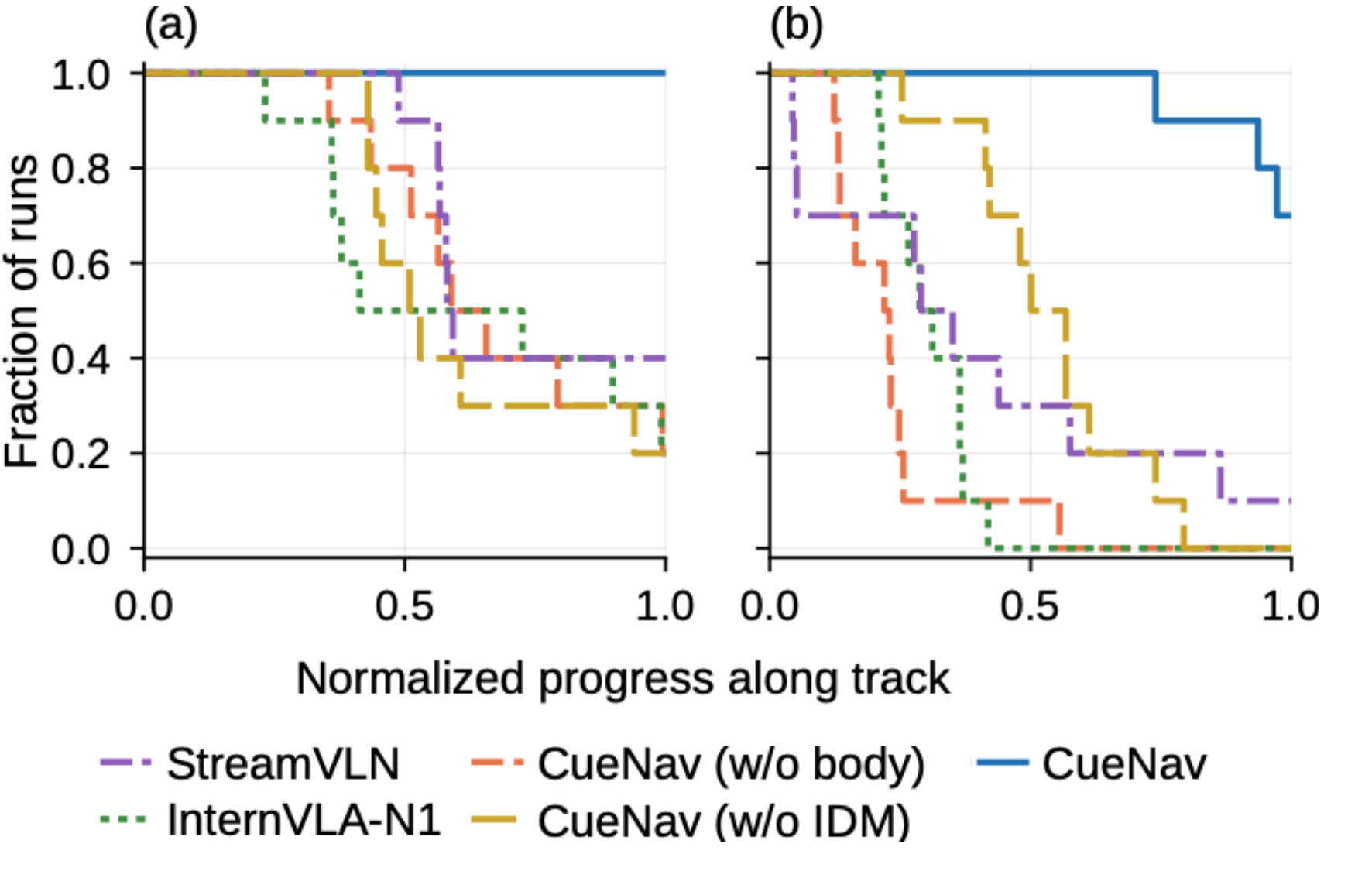}
    \caption{Normalized progress along the narrow-track task over 10 runs per method. Each curve shows the fraction of runs reaching at least a given progress. (a)~\qty{1.5}{\meter} track width. (b)~\qty{1.0}{\meter} track width.}
   \label{fig:precisenavigation_success}\vspace{-5mm}
\end{figure}

\textbf{Results.}
Fig.~\ref{fig:precisenavigation_success} shows that \textit{CueNav} maintains high progress as the track becomes more constrained. It completes all trials in the \qty{1.5}{\meter} setting and \qty{70}{\percent} of trials in the \qty{1.0}{\meter} setting, while the comparison methods frequently terminate before reaching the end. The ablations indicate that both the embodiment-visible observation and the proposed \ac{idm} contribute to precise navigation. Removing the embodiment cue degrades performance in the narrower passage, while replacing the \ac{idm} with geometry-based grounding also reduces progress despite using the same video planner and visual observation. Fig.~\ref{fig:narrow_gap_1m} provides a qualitative example of \textit{CueNav}'s generated visual plan and the corresponding real observations during navigation. Fig.~\ref{fig:precisenavigation_pathoverlay} further illustrates the effect of action grounding. \textit{CueNav} closely realizes the generated visual plan, whereas \textit{CueNav w/o IDM} shows visible deviations between the generated and realized observations, indicating less accurate translation of the visual plan into robot motion.

\subsection{Planning beyond Local Guidance with Global Task Context}
\label{subsec:maze}
We use maze navigation as a controlled setting in which reaching a distant goal requires a sequence of decisions over multiple local observations. This allows us to evaluate whether providing global task context helps the video planner make longer-horizon navigation decisions.

\textbf{Simulation Environment and Task Setup.}
We use simulated maze environments in DeepMind Lab~\cite{beattie2016deepmind} with grid sizes ranging from $3\times3$ to $6\times6$. The agent is controlled by continuous forward linear velocity and yaw rate commands at \qty{10}{\hertz}. The agent's initial position and a red-box goal are randomly placed in the maze, and the agent navigates to the goal without receiving turn-by-turn directional commands.

\begin{figure*}[t]
  \centering
  \includegraphics[width=0.95\textwidth]{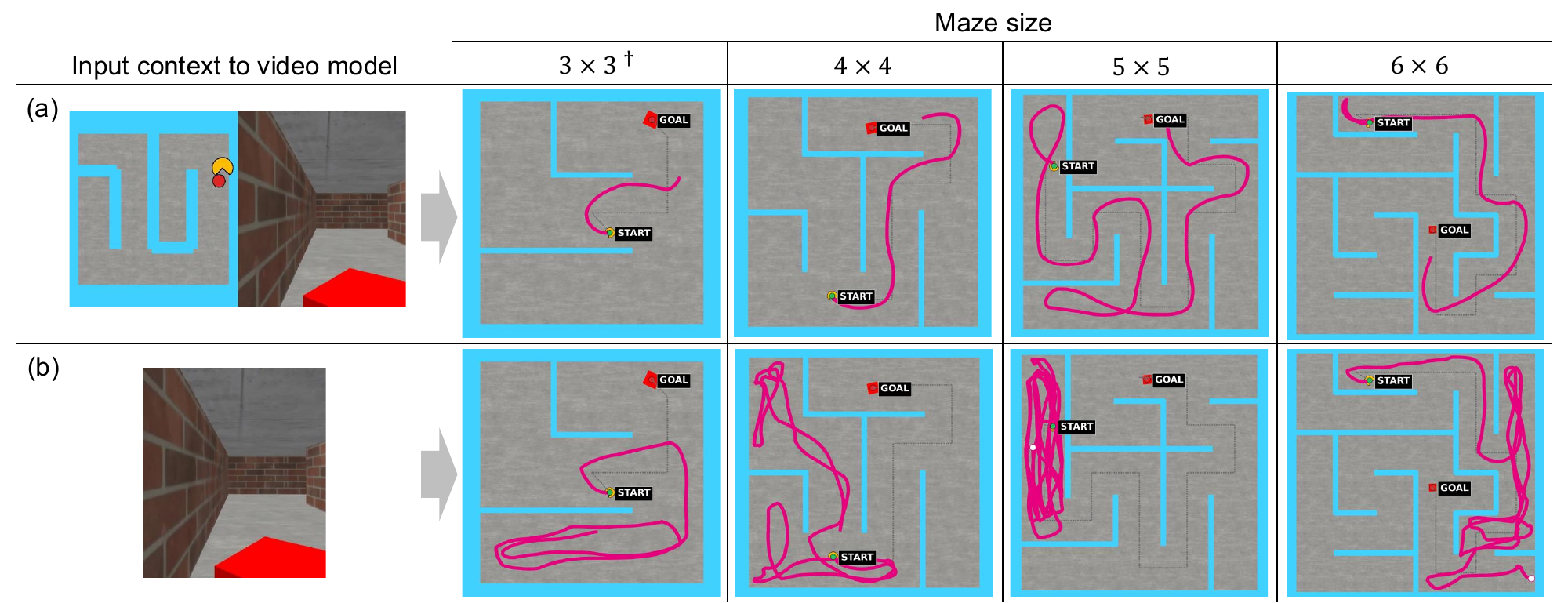}
  \caption{Maze navigation trajectories (magenta) with (a) global task context provided as a visual cue and (b) egocentric view only. Both variants are trained on $3\times3$ mazes ($\dagger$ denotes the training distribution) and evaluated zero-shot on larger mazes up to $6\times6$.}
  \label{fig:maze_trajectory}\vspace{-5mm}
\end{figure*}

The visual input to \textit{CueNav} combines the egocentric view with a \ac{bev} map indicating the agent and goal locations while leaving the solution path unspecified (See Fig.~\ref{fig:maze_trajectory}(a)). The two views are tiled into a single $832\times480$ pixel observation (width $\times$ height), providing the video planner with global task context in addition to the egocentric view. For action grounding, the \ac{idm} is trained on flow fields computed from the egocentric view portion of the observation, keeping the \ac{bev} cue specific to the video planner.

\textbf{Training and Evaluation Protocol.}
Training data is collected from 300 successful goal-reaching trajectories in randomly generated $3\times3$ mazes. In each maze, the agent follows a solution trajectory generated by a local waypoint controller, and the resulting observations are used to post-train the video planner. The planner is therefore exposed only to $3\times3$ mazes during post-training and is evaluated zero-shot on larger maze sizes up to $6\times6$. We evaluate 20 randomly generated episodes for each maze size and report the fraction of trials that reach the goal as a function of simulation time.

\textbf{Ablation on Global Task Context.}
To isolate the contribution of the visual cue providing global task context, we compare \textit{CueNav} with \textit{CueNav (w/o map)}, a variant that receives only the egocentric view without access to the \ac{bev} map, as shown in Fig.~\ref{fig:maze_trajectory}(b).

\begin{figure}[htpb]
   \centering
\includegraphics[width=0.43\textwidth]{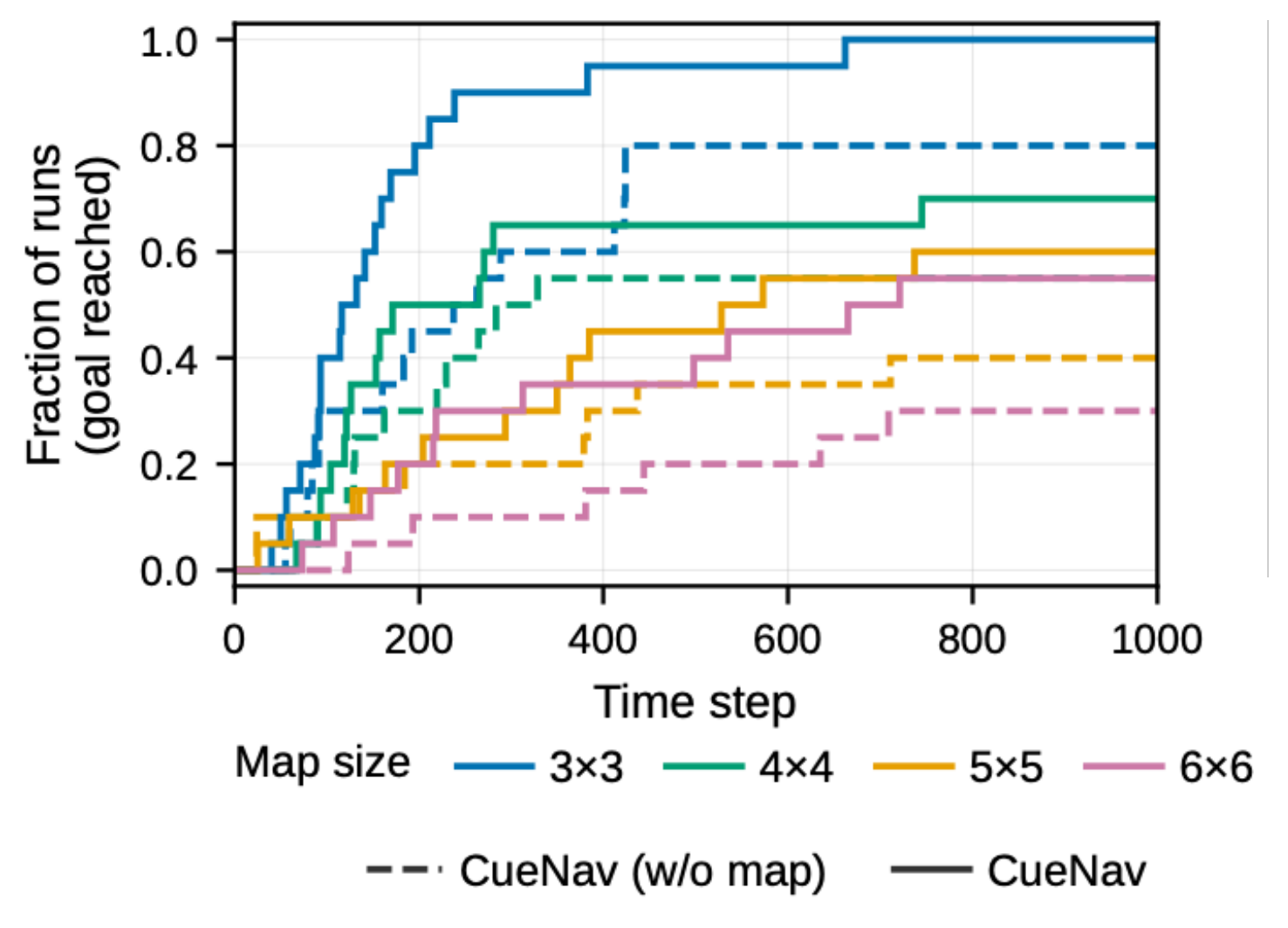}
   \caption{Goal-reaching performance across maze sizes.
Curves show the fraction of 20 runs that have reached the goal by each timestep.}
   \label{fig:maze_success} \vspace{-5mm}
\end{figure}

\textbf{Results.}
Fig.~\ref{fig:maze_success} shows that \textit{CueNav} with global task context consistently outperforms \textit{CueNav (w/o map)} across all maze sizes. The performance gap is already visible in the $3\times3$ training setting and becomes increasingly pronounced for larger $5\times5$ and $6\times6$ mazes, suggesting that global task context becomes more beneficial as navigation requires decisions over longer horizons. Although performance decreases with increasing maze size, \textit{CueNav} retains \qty{55}{\percent} success in unseen $6\times6$ mazes, compared with \qty{30}{\percent} for \textit{CueNav (w/o map)}.

Fig.~\ref{fig:maze_trajectory} further illustrates representative trajectories across increasing maze sizes. With the global map context, \textit{CueNav} produces more goal-directed trajectories, while \textit{CueNav (w/o map)} exhibits more local wandering and less consistent progress toward the distant goal. Together, these results indicate that providing global task context through visual observation enables planning beyond local directional guidance and supports zero-shot generalization to larger unseen environments.

\subsection{Case Study: Cross-embodiment Deployment with a Shared Video Planner}
\label{subsec:case_studies}
Beyond the controlled evaluations above, we examine cross-embodiment deployment by reusing the same video planner on the Unitree Go2 while training a separate \ac{idm} for the platform. Compared with the Husky A300, the Go2 has a different morphology and action space, while embodiment-specific control is handled by the platform-specific \ac{idm}.

\begin{figure*}[t]
  \centering
\includegraphics[width=0.99\textwidth]{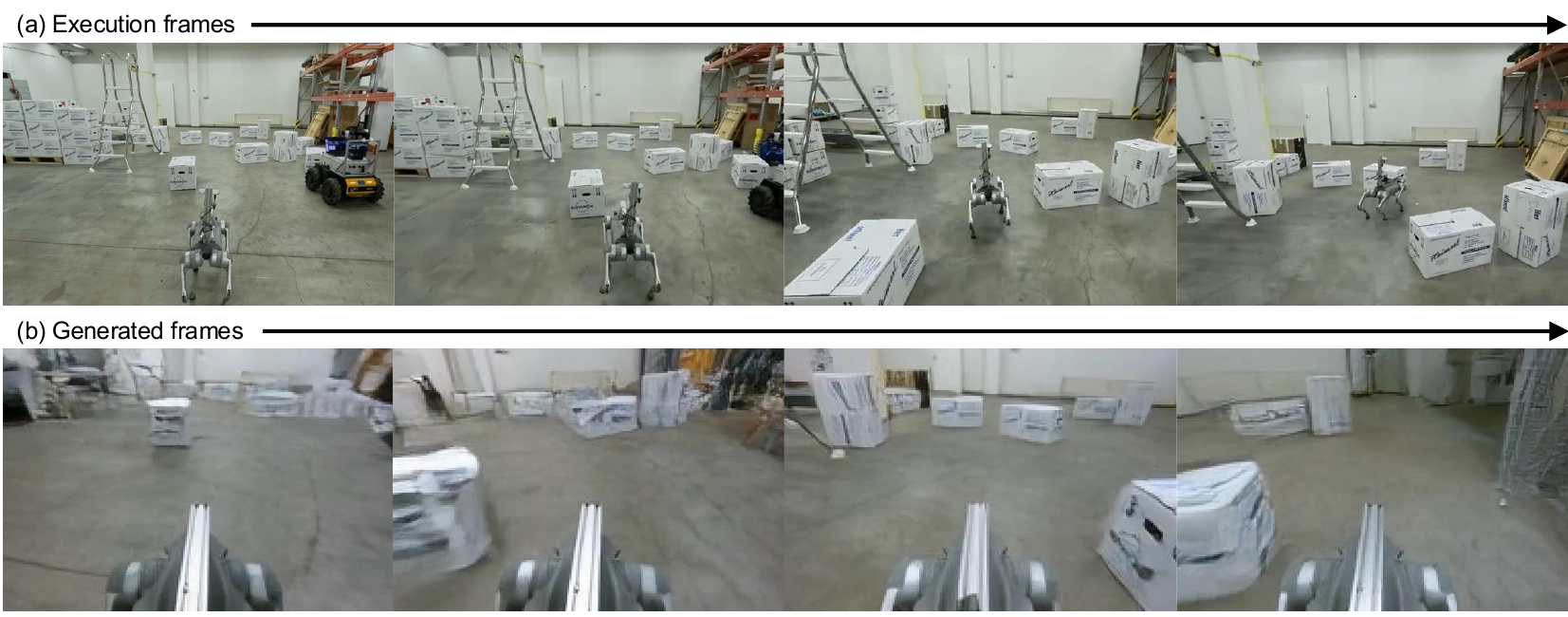}
\caption{Cross-embodiment deployment of CueNav on the Unitree Go2.
(a) Execution frames during zero-shot navigation in a cluttered environment.
(b) Corresponding generated future observations, demonstrating reuse of the shared video planner on a different robot embodiment.}
  \label{fig:embodiment_go2}\vspace{-3mm}
\end{figure*}

Fig.~\ref{fig:embodiment_go2} shows a zero-shot deployment on the Go2 in an environment not seen during post-training, where the robot traverses randomly placed obstacles without collision. The embodiment-visible observation provides the planner with visual information about the executing robot and its spatial relation to nearby obstacles, allowing the shared planner to generate behavior appropriate to the different embodiment. This case study illustrates that \textit{CueNav} can reuse a common video planner across embodiments while adapting execution through an embodiment-specific \ac{idm}.

\section{Conclusion and Future Work}
We presented CueNav, a video-based navigation framework that conditions a generative video planner on task- and embodiment-relevant visual cues and translates predicted visual motion into continuous robot actions using a flow-based \ac{idm}. Our results show that this formulation enables navigation beyond local directional guidance and precise embodiment-aware control while supporting generalizable navigation across different robot platforms. Despite these results, several limitations remain. First, video-planning inference remains computationally expensive, limiting the replanning rate during deployment. Few-step or latent-space distillation could help alleviate this bottleneck. Second, the current video model predicts only a short horizon and conditions on a finite observation window, limiting longer-horizon reasoning and memory. Longer-term visual context or hierarchical memory could extend the planner beyond its current temporal horizon. Finally, the \ac{idm} is tied to a fixed camera configuration, and its flow-to-action mapping may not transfer directly across different camera poses. Camera-conditioned or adaptive \acp{idm} could improve generalization across sensor configurations.

\section*{ACKNOWLEDGMENT}
\ifanon
\else
The authors gratefully acknowledge funding and computational resources provided by AMD through the MIT AI hardware program and the AMD University Program’s AI $\&$ HPC Cluster.
\fi
The authors acknowledge the use of ChatGPT (GPT-5.5) and Claude (Opus 5)
for language editing and figure/table formatting; the images in Fig.~1
were generated with ChatGPT. All AI-assisted content was verified by the
authors.  
\vspace{-3mm}  
$ $

\addtolength{\textheight}{-0cm}   






\bibliographystyle{IEEEtran}
\bibliography{ref.bib}

@article{beattie2016deepmind,
  title={Deepmind lab},
  author={Beattie, Charles and Leibo, Joel Z and Teplyashin, Denis and Ward, Tom and Wainwright, Marcus and K{\"u}ttler, Heinrich and Lefrancq, Andrew and Green, Simon and Vald{\'e}s, V{\'\i}ctor and Sadik, Amir and others},
  journal={arXiv preprint arXiv:1612.03801},
  year={2016}
}

@inproceedings{harley2025alltracker,
  title={Alltracker: Efficient dense point tracking at high resolution},
  author={Harley, Adam W and You, Yang and Sun, Xinglong and Zheng, Yang and Raghuraman, Nikhil and Gu, Yunqi and Liang, Sheldon and Chu, Wen-Hsuan and Dave, Achal and You, Suya and others},
  booktitle={2025 IEEE/CVF International Conference on Computer Vision (ICCV)},
  pages={5253--5262},
  year={2025},
  organization={IEEE}
}

@article{chen2024diffusion,
  title={Diffusion forcing: Next-token prediction meets full-sequence diffusion},
  author={Chen, Boyuan and Mart{\'\i} Mons{\'o}, Diego and Du, Yilun and Simchowitz, Max and Tedrake, Russ and Sitzmann, Vincent},
  journal={Advances in Neural Information Processing Systems},
  volume={37},
  pages={24081--24125},
  year={2024}
}

@article{li2026turning,
  title={Turning Video Models into Generalist Robot Policies},
  author={Li, Sizhe Lester and Kim, Evan and Bai, Xingjian and Zhao, Tong and Pang, Tao and Simchowitz, Max and Sitzmann, Vincent},
  journal={arXiv preprint arXiv:2605.27817},
  year={2026}
}

@article{chen2026imaginav,
  title={Imaginav: Scalable embodied navigation via generative visual prediction and inverse dynamics},
  author={Chen, Jie and Cai, Yuxin and Wang, Yizhuo and Bai, Ruofei and Cao, Yuhong and Li, Jun and Yun, Yau Wei and Sartoretti, Guillaume},
  journal={arXiv preprint arXiv:2603.13833},
  year={2026}
}

@article{zhang2026sparse,
  title={Sparse video generation propels real-world beyond-the-view vision-language navigation},
  author={Zhang, Hai and Liang, Siqi and Chen, Li and Li, Yuxian and Xu, Yukuan and Zhong, Yichao and Zhang, Fu and Li, Hongyang},
  journal={arXiv preprint arXiv:2602.05827},
  year={2026}
}

@inproceedings{wei2026ground,
  title={Ground slow, move fast: A dual-system foundation model for generalizable vision-language navigation},
  author={Wei, Meng and Wan, Chenyang and Peng, Peng and Yu, Xiqian and Yang, Yuqiang and Feng, Delin and Cai, Wenzhe and Zhu, Chenming and Wang, Tai and Pang, Jiangmiao and others},
  booktitle={International Conference on Learning Representations},
  volume={2026},
  pages={12380--12396},
  year={2026}
}

@article{wei2025streamvln,
  title={Streamvln: Streaming vision-and-language navigation via slowfast context modeling},
  author={Wei, Meng and Wan, Chenyang and Yu, Xiqian and Wang, Tai and Yang, Yuqiang and Mao, Xiaohan and Zhu, Chenming and Cai, Wenzhe and Wang, Hanqing and Chen, Yilun and others},
  journal={arXiv preprint arXiv:2507.05240},
  year={2025}
}

@article{majumdar2026robostral,
  title={Robostral Navigate},
  author={Majumdar, Arjun and Sooriyarachchi, Avinash and Tibi, Benjamin and Bamford, Chris and Chane-Sane, Elliot and Lample, Guillaume and Chandu, Khyathi Raghavi and Fuh, Ludovic Ho and Poiree, Mathieu and Duchenne, Olivier and others},
  journal={arXiv preprint arXiv:2607.20785},
  year={2026}
}

@article{gong2026abot,
  title={ABot-N1: Toward a General Visual Language Navigation Foundation Model},
  author={Gong, Ruiyan and Guo, Yingnan and Hu, Junjun and Kong, Jintao and Leng, Xiaoxu and Li, Tianlun and Li, Weize and Liu, Fei and Liu, Zhicheng and Lu, Jia and others},
  journal={arXiv preprint arXiv:2607.10383},
  year={2026}
}

@inproceedings{zhang2026embodied,
  title={Embodied navigation foundation model},
  author={Zhang, Jiazhao and Li, Anqi and Qi, Yunpeng and Li, Minghan and Liu, Jiahang and Wang, Shaoan and Liu, Haoran and Zhou, Gengze and Wu, Yuze and Li, Xingxing and others},
  booktitle={International Conference on Learning Representations},
  volume={2026},
  pages={127293--127322},
  year={2026}
}

@inproceedings{leroy2024grounding,
  title={Grounding image matching in 3d with mast3r},
  author={Leroy, Vincent and Cabon, Yohann and Revaud, J{\'e}r{\^o}me},
  booktitle={European conference on computer vision},
  pages={71--91},
  year={2024},
  organization={Springer}
}

@article{grover2025enhancing,
  title={Enhancing generalization in vision-language-action models by preserving pretrained representations},
  author={Grover, Shresth and Gopalkrishnan, Akshay and Ai, Bo and Christensen, Henrik I and Su, Hao and Li, Xuanlin},
  journal={arXiv preprint arXiv:2509.11417},
  year={2025}
}

@article{liu2026imagineuav,
  title={ImagineUAV: Aerial Vision-Language Navigation via World-Action Modeling and Kinodynamic Planning},
  author={Liu, Xuchen and Huang, Jiawei and Xia, Shihao and Liu, Bingxi and Cui, Jinqiang and Yang, Jiankun},
  journal={arXiv preprint arXiv:2606.01205},
  year={2026}
}

@article{wan2025wan,
  title={Wan: Open and advanced large-scale video generative models},
  author={Wan, Team and Wang, Ang and Ai, Baole and Wen, Bin and Mao, Chaojie and Xie, Chen-Wei and Chen, Di and Yu, Feiwu and Zhao, Haiming and Yang, Jianxiao and others},
  journal={arXiv preprint arXiv:2503.20314},
  year={2025}
}

@article{huang2026navdreamer,
  title={Navdreamer: Video models as zero-shot 3d navigators},
  author={Huang, Xijie and Gai, Weiqi and Wu, Tianyue and Wang, Congyu and Zheng, Qiaoyu and Liu, Zhiyang and Zhou, Xin and Wu, Yuze and Gao, Fei},
  journal={IEEE Robotics and Automation Letters},
  year={2026},
  publisher={IEEE}
}

@inproceedings{anderson2018vision,
  title={Vision-and-language navigation: Interpreting visually-grounded navigation instructions in real environments},
  author={Anderson, Peter and Wu, Qi and Teney, Damien and Bruce, Jake and Johnson, Mark and S{\"u}nderhauf, Niko and Reid, Ian and Gould, Stephen and Van Den Hengel, Anton},
  booktitle={2018 IEEE/CVF conference on computer vision and pattern recognition},
  pages={3674--3683},
  year={2018},
  organization={IEEE}
}

@inproceedings{yokoyama2024hm3d,
  title={Hm3d-ovon: A dataset and benchmark for open-vocabulary object goal navigation},
  author={Yokoyama, Naoki and Ramrakhya, Ram and Das, Abhishek and Batra, Dhruv and Ha, Sehoon},
  booktitle={2024 IEEE/RSJ International Conference on Intelligent Robots and Systems (IROS)},
  pages={5543--5550},
  year={2024},
  organization={IEEE}
}

@inproceedings{wang2025trackvla,
  title={Trackvla: Embodied visual tracking in the wild},
  author={Wang, Shaoan and Zhang, Jiazhao and Li, Minghan and Liu, Jiahang and Li, Anqi and Wu, Kui and Zhong, Fangwei and Yu, Junzhi and Zhang, Zhizheng and Wang, He},
  booktitle ={Proceedings of The 9th Conference on Robot Learning},
  year={2025},
  publisher={PMLR},
}

@article{hu2021lora,
  title={Lora: Low-rank adaptation of large language models},
  author={Hu, Edward J and Shen, Yelong and Wallis, Phillip and Allen-Zhu, Zeyuan and Li, Yuanzhi and Wang, Shean and Wang, Lu and Chen, Weizhu},
  journal={arXiv preprint arXiv:2106.09685},
  year={2021}
}

@inproceedings{liu2025x,
  title={X-mobility: End-to-end generalizable navigation via world modeling},
  author={Liu, Wei and Zhao, Huihua and Li, Chenran and Biswas, Joydeep and Okal, Billy and Goyal, Pulkit and Chang, Yan and Pouya, Soha},
  booktitle={2025 IEEE International Conference on Robotics and Automation (ICRA)},
  pages={7569--7576},
  year={2025},
  organization={IEEE}
}

@inproceedings{sridhar2024nomad,
  title={Nomad: Goal masked diffusion policies for navigation and exploration},
  author={Sridhar, Ajay and Shah, Dhruv and Glossop, Catherine and Levine, Sergey},
  booktitle={2024 IEEE International Conference on Robotics and Automation (ICRA)},
  pages={63--70},
  year={2024},
  organization={IEEE}
}

@article{serpiva2026dreamtonav,
  title={DreamToNav: Generalizable Navigation for Robots via Generative Video Planning},
  author={Serpiva, Valerii and Sam, Jeffrin and Simon, Chidera and Amjad, Hajira and Zhura, Iana and Lykov, Artem and Tsetserukou, Dzmitry},
  journal={arXiv preprint arXiv:2603.06190},
  year={2026}
}

\end{document}